%% file: main.tex
\documentclass{bmvc2k}
\newcommand{\cmark}{\ding{51}}%
\newcommand{\xmark}{\ding{55}}%
\usepackage{graphicx}
\usepackage{comment}
\usepackage{textcomp}
\usepackage{graphicx}
\usepackage{siunitx}
\usepackage{tabularx}

\usepackage{amsmath,amssymb}
\usepackage{multirow}
\usepackage{amssymb}
\usepackage{booktabs}
\usepackage{amssymb}
\usepackage{pifont}
\usepackage{float}
\usepackage{makecell}
\usepackage{dirtytalk}
\newcommand{\methodName}{KWS-Net}
\title{Seeing wake words:\\ Audio-Visual Keyword Spotting}

\addauthor{Liliane Momeni}{liliane@robots.ox.ac.uk}{1}
\addauthor{Triantafyllos Afouras}{afourast@robots.ox.ac.uk}{1}
\addauthor{Themos Stafylakis}{tstafylakis@omilia.com}{2}
\addauthor{Samuel Albanie}{albanie@robots.ox.ac.uk}{1}
\addauthor{Andrew Zisserman}{az@robots.ox.ac.uk}{1}

\addinstitution{
 Visual Geometry Group\\
 Department of Engineering Science\\
 University of Oxford\\
 Oxford, UK
}
\addinstitution{
 Omilia Conversational Intelligence\\
 Athens, Greece\\
}

\runninghead{Momeni, Afouras, Stafylakis, Albanie, Zisserman}{Audio-visual KWS}

\newcommand{\lili}[1]{\textcolor{black}{#1}}

\begin{document}

\maketitle

\begin{abstract}
The goal of this work is to automatically determine
\textit{whether} and \textit{when} a word of interest is spoken by a talking face, with or without
the audio. We propose a  \textit{zero-shot} method suitable for `\textit{in the wild}' videos.
Our key contributions are: 
(1)~a novel convolutional architecture, KWS-Net, that uses a \textit{similarity map} intermediate representation to separate the task into (i)~\textit{sequence matching}, and (ii)~\textit{pattern detection}, to decide whether the word is there and when;
 (2)~we demonstrate that if audio is available, visual keyword spotting improves the performance both for a clean and noisy audio signal. Finally, (3)~we show that our method generalises to other languages, specifically French and German, and achieves a comparable performance to English with less language
specific data, by fine-tuning the network pre-trained on English.
The method exceeds the performance of the previous state-of-the-art visual keyword
spotting architecture when trained and tested on the same benchmark, and also
that of a state-of-the-art lip reading method.
\end{abstract}

\section{Introduction}
\label{sec:intro}

Keyword spotting (KWS) is the task of detecting a word of interest within continuous speech. In audio-visual data, the keyword can be detected from the audio stream only, from the visual stream only, or from both streams. The task differs from automatic speech recognition (ASR) or from automatic visual speech recognition (AVSR, lip reading),  where the aim is to recognise the phrases and sentences being spoken from scratch. In KWS, the word that is sought is provided by the user, and consequently the task is easier than recognising with no knowledge as in ASR or AVSR. This suggests that a KWS model can (i) be much simpler than ASR or AVSR, and (ii) have higher performance.  

\begin{figure}
    \centering
    \includegraphics[width=\textwidth]{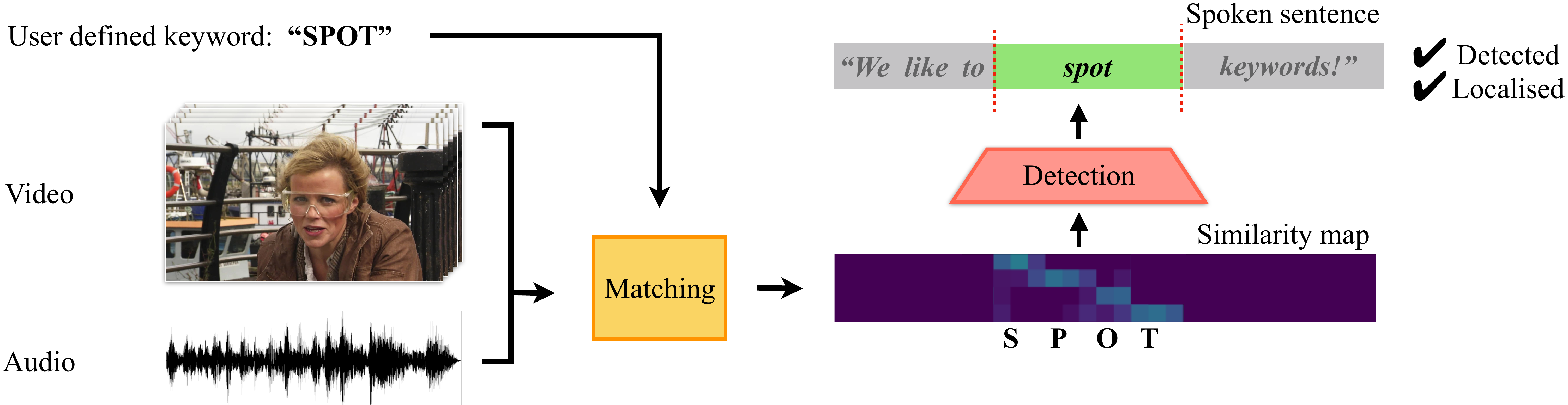}
   \vspace{-6mm}
    \caption{\textbf{General approach of \methodName}: The inputs to the model are a user-specified keyword and either audio, video or both. The objective is to detect \textit{whether} the keyword occurs in the input signal and, 
if present, then \textit{where} it is.}
    \label{fig:novel_architecture_main}
    \vspace{-5mm}
\end{figure}
KWS is more practical in many situations. Indeed, ASR is frequently not the aim of real-world speech processing applications and complete speech transcription can therefore be redundant. Keyword search, which consists in retrieving speech utterances including a keyword from a large database, is often a more useful task. \lili{KWS also surpasses ASR in cases where context is limited, for example for detecting mouthings in sign language~\cite{Albanie2020bsl1k}. }

Visual KWS has clear applications to cases where audio is unavailable such as for browsing archival silent films, and more importantly for cases where audio has been corrupted with noise, including for wake-word recognition (e.g.\ `OK Google', `Hey Siri' and `Alexa') as
 well as other human-robot interactions, such as in smart home
 technologies (for example, turning off the lights) or to assist people with speech impairment or aphonia \cite{Shillingford18}. 

A fundamental constraint for any visual KWS system
is detecting words which sound different but involve the same lip
movements (they have the same `visemes' -- visemes are the visual equivalent of phonemes; phonemes are the smallest unit of sound in speech).
For instance,
the words `may', `pay' and `bay' cannot be distinguished without audio as the visemes for `m', `p' and `b' look the same.
Other difficulties include intra-class differences (such
as accents, speed of speech and mumbling which modify lip movements)
and variable imaging conditions (such as lighting, motion,
resolution) \cite{Chung16}. Spotting words from continuous
speech is also challenging as there may be co-articulation of the lips.
 
In this paper, we introduce a novel convolutional architecture, \methodName, for spotting keywords in {\em visual} speech.
The model introduces a {\em similarity map} that splits the task into (i)~{\em matching} a token phoneme sequence against a viseme sequence, and (ii) \textit{detecting} an \textit{alignment pattern }to decide whether and when the keyword occurs (see Figure~\ref{fig:novel_architecture_main}).  Step~(ii)~is performed in a {\em detector-by-classification} manner, inspired by sliding window object detection methods. The model
is able to spot words that are \textit{unseen} during training, and are specified by a user at test time (zero-shot). We show that KWS-Net
exceeds the previous state-of-the-art network of Stafylakis\textit{ et al.}~\cite{themoskws} for visual KWS on standard benchmarks. Furthermore, we show that audio-visual KWS outperforms the audio-only KWS counterpart marginally for clean audio, but substantially for noisy audio. The visual-only and audio-visual KWS models are described in Section~\ref{sec:model}. Finally, we apply our method to French and German datasets built from TED videos (see Section~\ref{sec:exp_setup}) and  demonstrate that our model can perform comparably to English in other languages with less language specific training data. \lili{The project webpage is at: \small{\url{www.robots.ox.ac.uk/~vgg/research/kws-net/}}.}

\vspace{-3mm}
\section{Related Work} \label{sec:related_works}

\noindent{\textbf{Lip reading.}} Recent deep learning methods involving character-level recognition of visual sequences can be divided into two types: (i) models trained with a Connectionist Temporal Classification (CTC) loss \cite{Graves06}, where frame-wise label predictions are made in search for an optimal alignment with the output sequence, and (ii) models trained with a sequence-to-sequence (seq2seq) loss, that first read the entire input before attending to different parts of it at each step of an autoregressive output sequence prediction process. Examples of CTC models include LipNet~\cite{Assael16} and more recently LSVSR~\cite{Shillingford18}, that shows state-of-the-art performance with a word error rate as low as 40.9\% when trained on vast amounts of data. 
Examples of seq2seq models include the LSTM with attention model from Chung \textit{et al}.\ \cite{Chung17}, which extends the audio model `Listen, attend and spell'~\cite{chan2016listen} to visual and audio-visual ASR. Afouras \textit{et al}.\ \cite{Afouras18b} combine the seq2seq loss with self-attention layers and propose a transformer-based model. Hybrid approaches combining CTC and seq2seq losses were also recently proposed \cite{petridis2018audio, Afouras19}, demonstrating promising results on the LRS2 benchmark~\cite{Chung16b, Afouras19}.

\noindent{\textbf{Audio KWS.}} Traditional audio-based KWS methods are based on HMMs \cite{szoke2005comparison}.  
More recent deep learning works investigate fully connected networks \cite{Chen, Tucker}, time delay neural networks \cite{Myer2018EfficientKS, Sun17}, convolutional neural networks (CNNs) \cite{Sainath2015ConvolutionalNN,HelloEdge,Wang2, Palaz}, graph convolutional neural networks \cite{graph_cnn}, and recurrent neural networks (RNNs) \cite{Fernandez, HWang, Sun2016MaxpoolingLT}. RNNs are also combined with convolutional layers \cite{Arik, Lengerich, Kim} to simultaneously model local features and temporal dependencies. Recent works also explore seq2seq models for KWS \cite{Zhang18,Audhkhasi,Zhuang2016UnrestrictedVK,rosenberg2017end}.

\noindent\textbf{Visual KWS.} Yao \textit{et al}.~\cite{sliding-windows} use sliding windows to split sentence-level videos into smaller segments on which they perform word-level classification and aggregate across segments using a max pooling layer.  Their method is used for a closed-set of 1000 Mandarin keywords, whereas our method is zero-shot. We cannot compare to their work as (i) we do not have access to Mandarin phonetic dictionaries,  and (ii) their validation and test sets are unavailable. 
Jha \textit{et al}.\ \cite{Jha} propose a query by example visual KWS architecture, where the word query and retrieval are both videos, and a cosine similarity score is used to assign a label query to a target video. Recently, Stafylakis \textit{et al}.\ \cite{themoskws} devised an end-to-end architecture which uses RNNs to learn correlations between visual features and a keyword representation, extracted from a grapheme-to-phoneme encoder-decoder.

\noindent\textbf{Audio-visual KWS.} Ding \textit{et al}.\ \cite{Ding} build an audio-visual decision fusion KWS system, consisting of 2D CNNs to model the time-frequency features of the log mel-spectrogram and 3D CNNs to model the spatio-temporal features of the mouth. The softmax outputs of the audio and visual networks are combined through a summation, with fixed weights for each modality, to estimate the posterior probability of each keyword. In \cite{7389408}, adaptive decision audio-visual fusion based on HMMs is performed using a proposed lip descriptor. Both of these works are evaluated on the private, relatively small PKU-AV dataset of 3000 clips and 30 keywords, involving no more than 20 speakers and excluding any mouth occlusions. These methods are evaluated with keywords seen during training, as opposed to zero-shot.

\begin{figure}[h!]
  \vspace{-8mm}
    \centering
    \includegraphics[width=\textwidth]{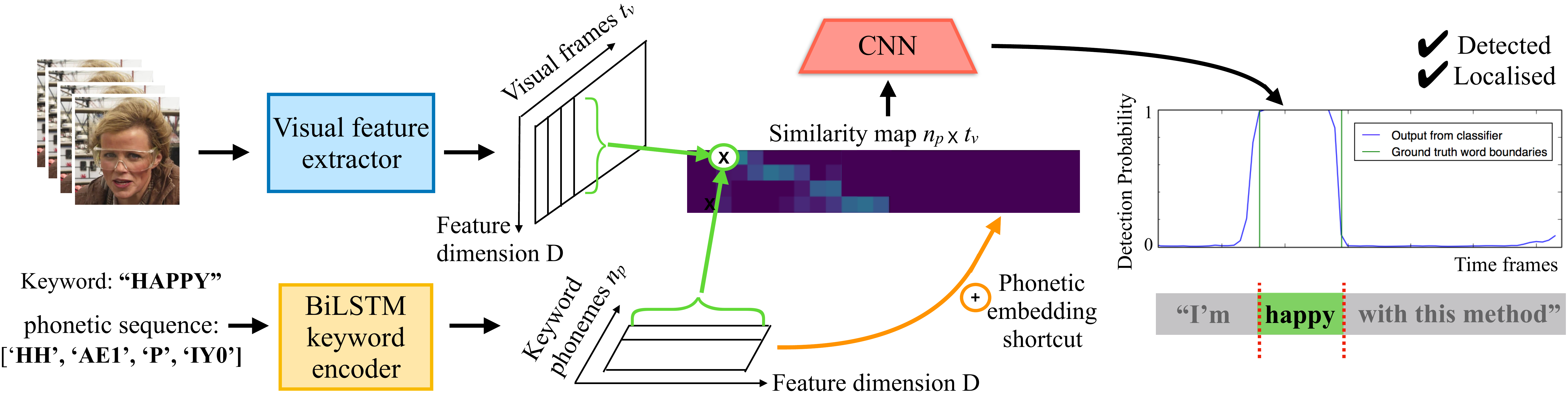}
     \vspace{-8mm}
    \caption{\textbf{Visual-only \methodName{} pipeline}: The viseme  and phonetic sequence embeddings are used to compute a similarity map, which is expected to show a strong diagonal component when the keyword is present. This pattern can be detected by a CNN-based classifier. The output keyword detection probabilities are plotted for the clip. See details in Section~\ref{sec:model}.}
    \label{fig:visual_only}
    \vspace{-7mm}
\end{figure}

\section{KWS-Net} \label{sec:model}

\noindent The visual KWS-Net model, shown in Figure~\ref{fig:visual_only}, \lili{contains two input streams: a visual feature extractor and a keyword encoder that produces an embedding for the pronunciation of the queried keyword. The visual and phonetic representations are fused into a single channel similarity bottleneck, which is then passed through a CNN classifier to detect an alignment pattern.}
Full details of the model are given in the appendix.

\noindent \textbf{Visual feature extractor.} The visual feature extractor takes as input a sequence of frames from a clip of a talking face and outputs visual features. The feature extraction is based on an 18-layer spatio-temporal ResNet~\cite{He15,themos2} 
which has shown good results on related tasks such as lip reading \cite{Afouras19} and audio-visual speech enhancement \cite{Afouras18}.
The network applies 3D convolutions on the input image sequence, followed by a 2D ResNet that gradually decreases the spatial dimensions, while preserving the temporal resolution. The visual encoding obtained is a sequence of dimension $t_v\times 512$, where $t_v$ is the number of input frames. The features are then passed through a BiLSTM \cite{Hochreiter97,lilischuster} to model temporal dynamics.

\noindent  \textbf{Keyword encoder.} The keyword encoder is a BiLSTM that ingests the phoneme token sequence of the input keyword (e.g.\ `HH,' `AE1,' `P,' `IY0' for `happy'), obtained using the CMU pronouncing dictionary \cite{cmu}, and outputs a phonetic keyword embedding sequence with dimensions $n_p\times512$, where $n_p$ is the number of phonemes in the keyword.

\noindent  \textbf{Similarity map.} We compute the \lili{dot product} between the phonetic sequence embedding $P$ ($n_p\times512$) and the visual feature sequence $V$ ($t_v\times 512$) which results in a similarity map ($n_p\times t_v$), expected to show high activation when the keyword occurs in the clip (positive pair), i.e. when the two modalities align.

\noindent  \textbf{CNN detector and classifier.} The similarity map is processed by a shallow CNN, which outputs the probability that the keyword is present at a specific location, by detecting patterns in it (e.g.\ a strong diagonal component).
The CNN gradually subsamples the temporal dimension by a factor of 8 and collapses the phoneme dimension to a singleton, \lili{resulting in an output of length $t_v^{out} = t_v / 8$ .} We apply a sigmoid activation on the resulting temporal sequence that outputs for every frame the probability that the keyword occurs around it.
The sample is predicted to contain the keyword if the maximum probability over all the frames is above a certain threshold, and the frame position of the maximum is regarded as the predicted location of the keyword. 

As shown in Figure~\ref{fig:visual_only}, before feeding the similarity map to the CNN, we concatenate the phonetic sequence embedding (broadcast over time) to it. The intuition for the addition of this shortcut is the following: (i) Some phonemes have a short duration so they may not appear in the map, especially in visual-only experiments where the frame rate is 25Hz. (ii)~Some phonemes may appear more than once in the keyword, meaning the diagonal assumption of the pattern might no longer hold since off-diagonal components may appear.

\noindent \textbf{Loss function.} For training we create clip-keyword sample pairs which are labeled positive or negative
depending on whether or not the keyword occurs in the clip (which can contain an arbitrarily long utterance). \lili{
Given a sample pair, the \methodName{} model outputs a probability $p_t(y=1|V,P)$ representing how likely the keyword is to occur at every temporal location $t \in [1, t_v^{out}]$.
We obtain a sequence-level prediction by taking the maximum probability over all time locations. 
The optimisation objective is then a binary cross-entropy loss between this prediction and the ground truth sequence-level label $y^*$ (1 for positive sample, 0 otherwise):} 

\lili{
\begin{equation}\label{eq:1}
L_{kws}(V, P, y^*) = - y^* \ log \max_t{p_t(y=1|V,P)} - (1-y^*) \ log (1- \max_t{p_t(y=1|V,P)} ) 
\end{equation}
}
If we have access to the exact word time boundaries then the temporal interval is used as extra supervision to help the model learn to correctly localise keywords within the clip: for positive samples, we calculate the maximum only within those time boundaries where the keyword is known to occur, \lili{ instead of the full length $[1, t_v^{out}]$.} If not stated otherwise, this is the method that we use.
The boundaries can be obtained by forced alignment and are included with some datasets (e.g.~LRS2 \cite{Chung16b}). 

\noindent \textbf{Differences to prior work.} Here, KWS is converted to an object detection problem where the CNN detects patterns from a similarity map that correspond to alignments between viseme and phonetic sequences. Similar alignments can be detected by word-level HMMs, that typically follow a `left-to-right, no skips' structure. \lili{Instead of detecting these patterns with probabilistic models, we employ a CNN and train the whole architecture jointly in an end-to-end manner, leveraging the large size of the datasets (see Table~\ref{table:datasets}) and following the recent trend in lip reading state-of-the-art methods (see Section~\ref{sec:related_works}). }

In~\cite{themoskws}, fixed length word embeddings are obtained from a grapheme (character) to phoneme (G2P) encoder-decoder architecture, using an additional decoder loss to encourage word representations that reflect the pronunciation. Instead, we build variable length word embeddings by directly encoding the phonemes using simply a BiLSTM. 
\noindent This approach has several advantages: (i) it strongly reflects the pronunciation and aligns better with the viseme features, (ii) it offers more control of words with multiple pronunciations, compared to G2P, and finally, (iii) phonemes are more language-independent compared to graphemes, enabling the encoder to be shared between languages.

\noindent \textbf{Audio-only KWS-Net. }We design an audio-only variation of the model, that operates on audio waveforms instead of video clips. We extract acoustic features by applying a STFT to the audio clip, with a 32ms window and 10ms hop-length, at a 16 kHz sample rate. The resulting spectrograms are projected to mel-scale, yielding 80-dimensional features. Since the video is sampled at 25 fps (40 ms per frame), every video input frame corresponds to 4 acoustic feature frames. The spectrograms are therefore passed through two strided convolutions to get the acoustic features down to video resolution, achieving a common temporal-scale for both modalities. This subsampling step allows us to keep the overall architecture the same for visual-only, audio-only and audio-visual inputs. 

\noindent \textbf{Audio-visual KWS-Net.} We employ a late decision audio-visual fusion. In this case, the audio-only and visual-only KWS-Net models are trained separately as explained above. The logits from the output of the CNN classifier from each of the audio-only and visual-only models are then averaged before applying the sigmoid activation, with the weights for each modality chosen according to the best performing value on the validation set. \lili{We explore the effect of varying modality weights in the appendix.}

   \vspace{-3mm}
\section{Experiments} \label{sec:exp_setup}

\noindent \textbf{Datasets.} The audio-visual datasets used are summarised in Table~\ref{table:datasets}. LRW~\cite{Chung16} consists of single-word utterances from BBC television broadcasts.  LRS2~\cite{Chung16b, Afouras19} and LRS3~\cite{Afouras18d} consist of thousands of spoken sentences from BBC and TED/TEDx talks respectively. \lili{Both datasets contain samples from multiple viewpoints, however} LRS3 is more challenging than LRS2: speakers are pictured from a wider range of viewpoints and with microphones/headsets, while addressing the audience results in more frequent head movements.  
\lili{We also use the French and German subsets of LRS3-Lang\footnote{Available at \url{www.robots.ox.ac.uk/~vgg/data/lip_reading/lrs3-lang}},  collected from TED/TEDx videos following the procedure from~\cite{Afouras18d}, and refer to them as LRS3-Fr and LRS3-De respectively.} In Section~\ref{sec:results}, we compare the performance of KWS-Net on LRS3-Fr and LRS3-De with that of LRS3, instead of LRS2, as the datasets come from the same domain. 

We set up our experiments following \cite{themoskws}:
for both training and evaluation, we use only keywords pronounced with $n_p\geqslant 6$ phonemes.
Moreover, as we want to evaluate on unseen keywords, we ensure that training and testing are performed on disjoint keyword vocabularies. 
\lili{To that end, we use all the words appearing in the test sets with $n_p\geqslant 6$ phonemes as evaluation keywords and we remove them from the training vocabulary, i.e. those words are not used in training the keyword encoder.}
We perform the language generalisation experiments on LRS3, LRS3-Fr, and LRS3-De in the seen and unseen keywords setting, therefore we drop the last constraint: test keywords for these datasets may have been seen during training. 
For exact details about the size of the resulting train and test keyword vocabulary of every dataset, please refer to the appendix.

\begin{table}
\footnotesize
\setlength{\tabcolsep}{6pt}
 \centering
\begin{tabularx}{\linewidth}{c c r r r r c}
\toprule
Dataset & Split & \#Utt. & \#Words & \#Hours  & Vocab. & Examples\\ 

\midrule
 \multirow{2}{*}{LRW~\cite{Chung16a}}    & Train-val & 514k & 514k&165 &500 & \multirow{2}{*}{\includegraphics[scale =0.13]{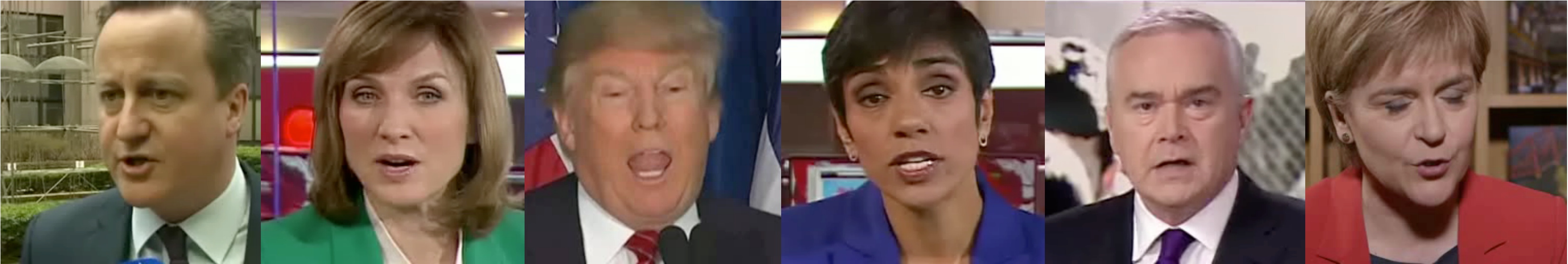}}\\
                      & Test      & 25k  &  25k& 8 &500 \\[0.1cm]
\midrule
 & Pre-train & 96k & 2M &195 & 41k & \multirow{3}{*}{\includegraphics[scale =0.13]{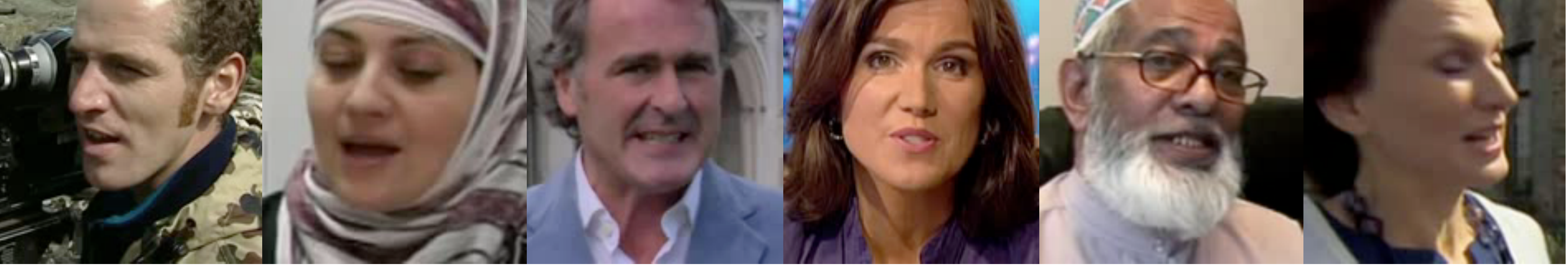}}\\
LRS2~\cite{Chung16b, Afouras19} & Train-val & 47k & 336k &29 & 18k \\
 & Test      & 1.2k & 6k &0.5 & 1.7k \\
 \midrule
 & Pre-train & 132k & 3.9M	& 444 & 51k & \multirow{3}{*}{\includegraphics[scale=0.13]{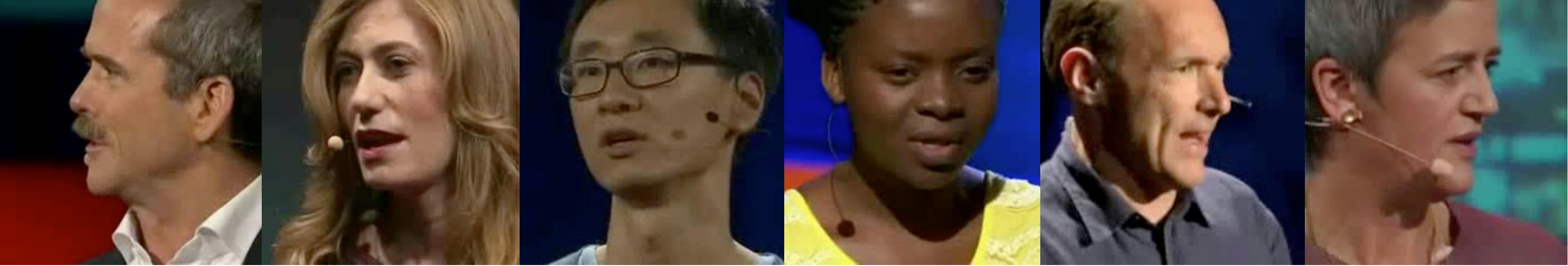}}\\
LRS3~\cite{Afouras18d} & Train-val & 32k & 358k&30 & 17k\\
 & Test      & 1.3k & 10k &1 &2k \\
 \midrule
  \multirow{2}{*}{LRS3-Fr}    & Train-val & 69k & 1M &107 & 28k &\multirow{2}{*}{\includegraphics[scale =0.13]{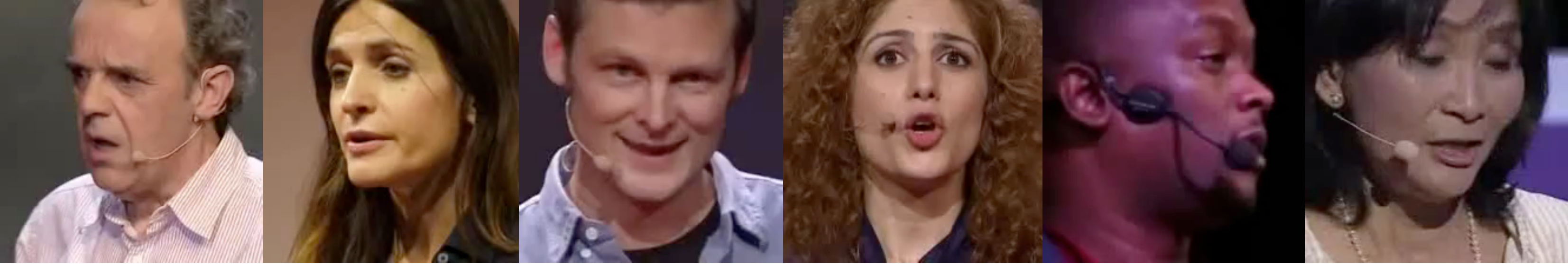}}\\
                      & Test      & 1.3k & 10.7k &1.2 & 2.1k  \\[0.1cm]
\midrule
  \multirow{2}{*}{LRS3-De}    & Train-val & 12k & 185k &20 & 11.2k &\multirow{2}{*}{\includegraphics[scale =0.13]{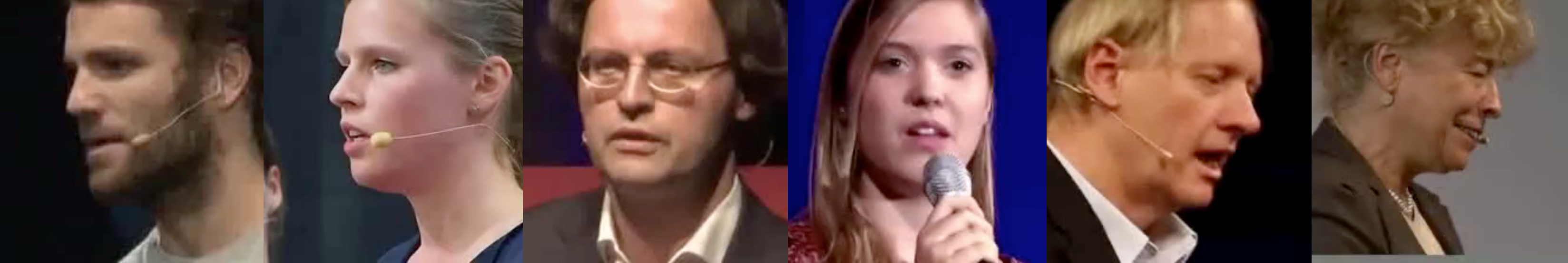}}\\
                      & Test  & 1.7k &10.6k &1.5 & 1.9k \\[0.1cm]
\bottomrule
\end{tabularx}

 \caption{\textbf{Statistics on datasets}: Division of development and test data, number of utterances and word instances, duration, vocabulary size and examples for each dataset.}
 \label{table:datasets}
\vspace{-5mm}
\end{table}

\noindent \textbf{Baselines.} We have four baselines: three are evaluated on LRS2 and the final one on LRW. \lili{As a first baseline we use our implementation of the model of Stafylakis \textit{et al}.\ \cite{themoskws}, which we also pre-train on LRW for fair comparison. This architecture is described fully in the appendix. Our second baseline is a variant of Stafylakis \textit{et al}.\ \cite{themoskws}, where the G2P network is switched to 
phoneme-to-grapheme (P2G) for a more expressive phonetic word representation.}

\lili{Our third baseline is the lip reading visual-ASR model from Afouras \textit{et al}.~\cite{Afouras2020ASRIA}, a CTC based model learned through cross-modal distillation, which is currently the state of the art on LRS2 for training only on publicly available data. The implementation code and pre-trained models are obtained from the authors. In order to apply the ASR model to KWS, we follow the method in~\cite{he2017streaming}: rather than only using the best decoding prediction, we extract the $n$ highest scoring hypotheses using a beam search and estimate the posterior probability that the keyword occurs in a clip using Equation (7) in~\cite{he2017streaming}.}

Our final baseline is the work of Jha \textit{et al}.\ \cite{Jha}, although our methods are not directly comparable as they perform query by example (as opposed to query by string). Their retrieval pipeline uses the LRW test set for querying and the LRW validation set for retrieval over 500 words. It should be noted that their model only works for a closed set of words, for which examples are provided, whereas KWS-Net can be used to spot words unseen during training. We directly compare to the results reported in their paper. 

\noindent \textbf{Ablations.} We consider three ablations for our visual-only \methodName{} architecture: (i) not using the word time boundaries for training, which we refer to as `no LOC' since this training regime does not explicitly encourage the correct localisation of the keyword,  (ii) removing the shortcut phonetic embedding, which we refer to as `no SH', and (iii) switching the BiLSTM keyword encoder for a P2G encoder-decoder, which we refer to as `+P2G'.

\noindent
\textbf{Pre-training and fine-tuning.} We initialise the weights of the ResNet-18 visual feature extractor \cite{themos2} from a model pre-trained on word-level lip reading (code and weights publicly available from \cite{Afouras18b}).
This part of the network is kept frozen during training: following the practice of \cite{Afouras18b}, we pre-compute the features on the entire datasets, then train the rest of the model directly on them to accelerate training.
We employ a curriculum training procedure for the rest of the network that consists of two stages:
(i) it is initially trained on the training set of LRW. As LRW contains clips of single words, here the model is trained without word time boundaries, (ii) the model is then fine-tuned on the sequence-level datasets.

\noindent
\textbf{Test setup.}
The performance of the models is evaluated on the test set of every dataset, using as queries all the held out test words (see datasets).
We look for each query keyword in all the clips of the test set.
Note that there is no balancing of positive and negative clips during evaluation: there are one or a few positive clips for a given keyword and the rest are negatives. During testing, in order to obtain fine-grained localisation, we apply the CNN classifier with a stride of one.

\noindent \textbf{Evaluation metrics.}  The performance is evaluated based on ranking metrics. For every keyword in the test vocabulary, we record the percentage of the total clips containing it that appear in the first N retrieved results, with N=[1,5,10], this is the `Recall at N' (R@N). \lili{Note that, since several clips may contain a query word, the maximum R@1 is not 100\%.} The mean average precision (mAP) and equal error rate (EER) are also reported. For each keyword-clip pair, the match is considered correct if the keyword occurs in the clip and the maximum detection probability occurs between the ground truth keyword boundaries. For each experiment, the average and standard deviation of each metric is computed over the last 5 checkpoints once the model has converged (validation loss has not improved for 5 epochs). 

\noindent
\textbf{Audio noise addition.} To investigate the robustness of the audio-only and audio-visual models against loud environments, we train by adding babble noise to the audio 50\% of the time with signal-to-noise-ratio (SNR) of 0 dB. Babble noise (interference from people talking simultaneously) is commonly used for audio degradation in audio-visual speech recognition~\cite{Afouras19,7389408} as it is more challenging than other types of environmental noise \cite{Babble}.

\section{Results} \label{sec:results}

\subsection{Visual-only \methodName{}}

\begin{table}
\footnotesize
\centering

\begin{tabularx}{\linewidth}{ l c c c c c c  }
\toprule

& R@1 & R@5 & R@10 & mAP & EER \\
 \midrule
  \midrule

 Stafylakis \& Tzimiropoulos (G2P) \cite{themoskws}* &22.8 & 49.0& 59.1& 36.0 &  8.9\\
 Stafylakis \& Tzimiropoulos (P2G) &30.0& 53.7 &65.3 & 43.5 &6.3 \\
 Visual-ASR \cite{Afouras2020ASRIA} & \textbf{41.9}  & 53.6 & 54.5& 51.3& - \\

 \midrule
 \textbf{KWS-Net}  &37.9 $\pm$ 0.3 &\textbf{66.8} $\pm$ 0.6&\textbf{75.6} $\pm$ 0.5&53.9 $\pm$ 0.3 &\textbf{5.7} $\pm$ 0.2\\

\hspace {3mm} no LOC  &37.2 $\pm$ 0.8&65.1 $\pm$ 0.2&73.7 $\pm$ 0.3 &53.0 $\pm$ 0.6 & 6.9 $\pm$ 0.4 \\

\hspace {3mm} no SH &35.0 $\pm$ 0.5 &62.4 $\pm$ 0.4&72.7 $\pm$ 0.9 &50.4 $\pm$ 0.3 & 7.5 $\pm$ 0.4\\
\hspace {3mm} +P2G & 39.1 $\pm$ 0.3 &  66.2 $\pm$ 0.6& 75.1 $\pm$ 0.4& \textbf{54.3 }$\pm$ 0.3&  5.9 $\pm$ 0.3\\

\bottomrule
\end{tabularx}

\caption{\textbf{Visual-only results}: Performance of baselines, visual-only \methodName{},  and ablations on the LRS2 test set. *refers to our implementation of \cite{themoskws} and Stafylakis \textit{et al}.\ P2G refers to switching G2P to P2G. Visual-ASR denotes our lip reading baseline from ~\cite{Afouras2020ASRIA}.  \methodName{} refers to our architecture from Section~\ref{sec:model}. no LOC represents not using the keyword time boundaries for training; no SH denotes not concatenating the phonetic embedding shortcut; +P2G denotes using a P2G encoder-decoder instead of a BiLSTM keyword encoder.}
    \label{tab:results_models}
\vspace{-5mm}
\end{table}

\noindent
\textbf{Baselines.} 
\lili{As can be seen in Table~\ref{tab:results_models},  Stafylakis \textit{et al}.\ G2P \cite{themoskws}* performs worse than the P2G baseline we propose.} Compared to Stafylakis \textit{et al}.\ P2G, \methodName{} significantly improves R@1 from  30.0\% to 37.9\% and mAP from 43.5\% to 53.9\%, with the EER also decreasing from 6.3\% to 5.7\%.  

\lili{KWS-Net has a higher R@5 compared to the lip reading visual-ASR baseline (66.8\% vs.\ 53.6\%) and a higher mAP  (53.9\% vs.\ 51.3\%). In fact, over a third of the keywords do not appear at all in the $n$-best list. KWS-Net has the advantage of retrieving more clips containing a keyword by using a higher R@N. Visual-ASR has a slightly higher R@1 (41.9\% vs.\ 37.9\%), but the method benefits from context of surrounding words. } 

Next, we replicate the test setting from \cite{Jha} and calculate their metrics on LRW: we achieve (not shown on the table) a higher P@10 of 77.1\% compared to 65.2\% and a higher R@10 of 15.4\% compared to 13.0\% as well as a slightly higher mAP of 57.8\% compared to 57.0\%. See \cite{Jha} for P@10 and R@10 metric definitions; note that R@N is defined differently in their experiments compared to in our work. 

\noindent\textbf{Ablations.} In Table \ref{tab:results_models}, we assess the value of each component of the architecture. For example when using the keyword time boundaries during training (see loss description in Section~\ref{sec:model}), the EER is reduced from 6.9\% to 5.7\%; however even if our method is trained without this extra annotation, KWS-Net no LOC still outperforms the Stafylakis \textit{et al}.\ P2G baseline (37.2\% vs.\ 30.0\% R@1). Similarly, the value of the phonetic shortcut embedding is shown in the decrease from 7.5\% to 5.7\% EER. \lili{Finally, we carry out an ablation by replacing the BiLSTM (KWS-Net) with P2G (KWS-Net+P2G), and conclude that the ablation performs overall worse than the original BiLSTM.}

\begin{figure}[h!]
    \centering
    \includegraphics[width=\textwidth]{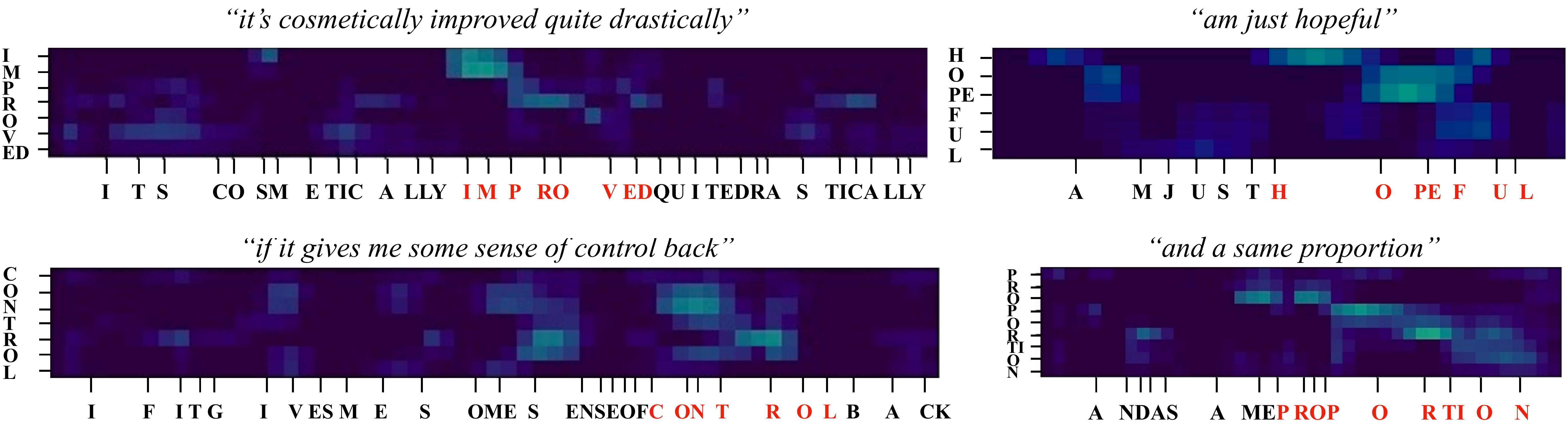}
   \vspace{-5mm}
    \caption{\textbf{Qualitative results}: Example similarity maps with visual-only KWS-Net for keywords `improved', `hopeful', `control' and `proportion' for clips in the LRS2 test set, with the application of a sigmoid for better visualisation. The vertical axis represents the phonemes in the keyword (graphemes are shown here for simplicity). The horizontal axis corresponds to the visual sequence; for visualisation we add phoneme ground truth start times for the entire clip utterance, with those corresponding to the keyword in red.}
    \label{fig:similarity_maps}
\end{figure}

\noindent\textbf{Visualisations.} \lili{In practice, we observe quasi-diagonal patterns in the similarity map visualisations in Figure~\ref{fig:similarity_maps}, which matches our intuition that viseme and phonetic feature sequences align when the keyword occurs in the clip. As explained in Section~\ref{sec:model}, there might be off-diagonal components due to repeated phonemes. Please refer to the appendix and project webpage for more qualitative examples.}
  \vspace{-5mm}

\noindent\textbf{Keyword length.} We explore how varying the minimum phoneme length of keywords $n_p$ effects the performance of visual-only \methodName{} on the LRS2 test set (see Table~\ref{tab:query_types}). As $n_p$ increases, the EER decreases and the mAP and R@1 increase as longer keywords are easier to visually spot. For this evaluation, additional shorter words are selected from the original LRS2 test set. Note, the network has not been trained for keywords with $n_p < 6$.  

\noindent \textbf{Phrases vs.\  Keywords.} We evaluate visual-only KWS-Net on the LRS2 test set, now using 3 word phrases as queries. For each of the evaluation unseen keywords, we construct a phrase query by concatenating the keyword with its preceding and succeeding words from the clip utterance, resulting in 666 phrases. The R@1 increases from 37.9\% to 65.3\% (see Table~\ref{tab:query_types}). 

\noindent \textbf{Seen vs.\  Unseen keywords.} We fine-tune our visual-only KWS-Net model, now including the previously unseen keywords from the LRS2 test set that occur in the training set (note there is no overlap between the training and testing videos). As seen in Table~\ref{tab:query_types}, the performance marginally improves for seen words compared to the zero-shot case, showing that our model is robust to words unseen during training (5.7\% vs.\ 5.1\% EER).  

\begin{table}

\centering

\footnotesize
\begin{tabularx}{\linewidth}{ l c c c c c c c  }
\toprule
query type &  $n_p$ & vocabulary & R@1 & R@5 & R@10& mAP & EER \\
 \midrule
  \midrule

unseen keywords &4 &1278&25.8 $\pm$ 0.4& 50.6 $\pm$ 0.5&61.2 $\pm$ 0.4&40.4 $\pm$ 0.3&11.5 $\pm$ 0.2\\
  unseen keywords& 6&644&37.9 $\pm$ 0.3& 66.8 $\pm$ 0.6&75.6 $\pm$ 0.5& 53.9 $\pm$ 0.3&5.7 $\pm$ 0.2\\
  unseen keywords& 8&227& 53.1 $\pm$ 0.9& 81.2 $\pm$ 0.5&87.1 $\pm$ 0.8& 68.9 $\pm$ 0.3&3.9 $\pm$ 0.4\\
 \midrule
seen keywords&6&644& 39.5 $\pm$ 0.6 &69.5 $\pm$ 0.4&78.9 $\pm$ 0.7 &56.7 $\pm$ 0.6&5.1 $\pm$ 0.2 \\
phrases & 9&666&65.3 $\pm$ 0.9&84.7 $\pm$ 0.3&89.1 $\pm$ 0.4&74.1 $\pm$ 0.6&3.7 $\pm$ 0.2\\ 
\bottomrule
\end{tabularx}

\caption{\textbf{Query investigation}: Performance of visual-only \methodName{} on the \textit{extended} LRS2 test set with different query types and minimum phoneme lengths $n_p$.}
\label{tab:query_types}
\vspace{-5mm}
\end{table}

\vspace{-5mm}
 \subsection{Audio-visual \methodName} 
We now look at whether we can augment audio with visual information. The results in Table~\ref{tab:audio_visual_main} indicate that lip movements improve performance  even when the audio signal is clean -- for example, R@1 increases from 67.7\% to 72.2\%. 
When the audio signal is corrupted with noise, the task of audio KWS becomes much harder. This is demonstrated by the decrease in R@1 from 67.7\% to 27.6\%. However, combining the audio and visual modalities results in a much higher performance, with R@1 increasing from 27.6\% to 52.7\%. The audio-visual model is more robust, surpassing the performance of both video-only and audio-only KWS-Net with a noisy audio signal, for a range of SNRs (-10 dB to 20 dB), as seen in Figure~\ref{tab:audio_visual_main}. 
\vspace{-8mm}
\begin{table}[!h]
   \vspace{-0pt}
	\begin{minipage}{0.55\linewidth}
		\centering
		\footnotesize
		\begin{tabular}{ c c c c c c c c }
    \toprule
    Mod. & Noise  & R@1 & R@5 & R@10& mAP & EER \\
     \midrule
      \midrule
    V & \xmark  &37.9&66.8&75.6&53.9&5.7\\
    A   &\xmark    &67.7&91.1& 94.6&83.3& 1.9\\
     AV & \xmark &72.2&94.7& 97.0&87.5& 1.7\\
    \midrule
    A & \cmark  &27.6 &49.8& 59.4 &39.7  & 12.8  \\
     AV & \cmark  &52.7 &81.9& 87.0  &69.6& 4.3 \\
    \bottomrule
    \end{tabular}
	\end{minipage}\hspace{5mm}
	\begin{minipage}{0.45\linewidth}
		\centering
		\includegraphics[width=50mm]{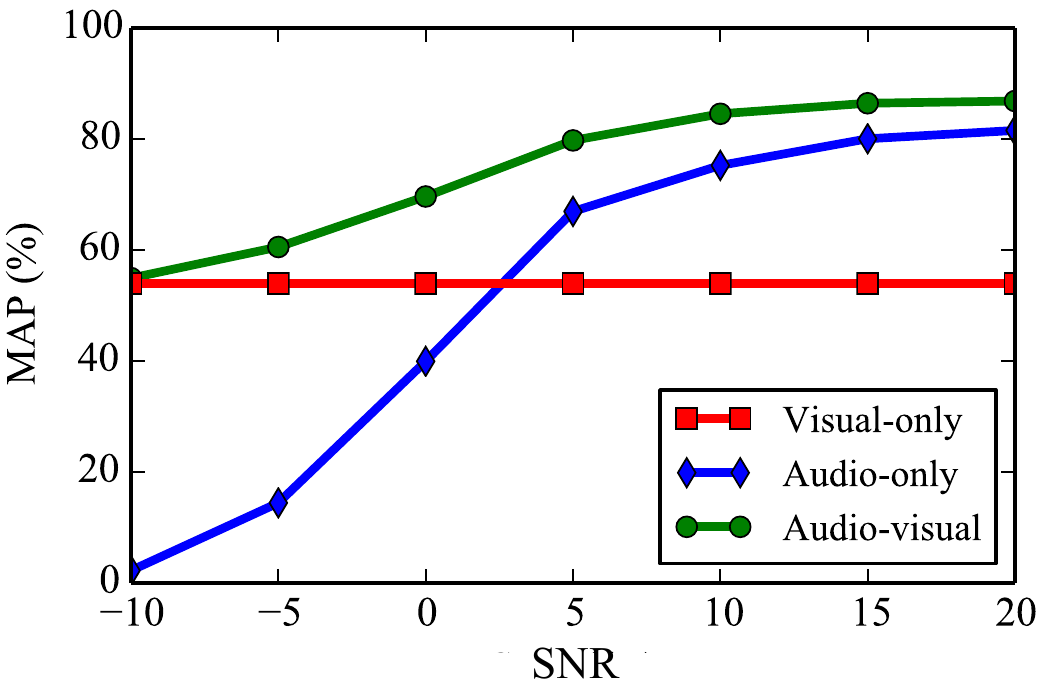}
		\label{ }
	\end{minipage}

\caption{(Left) \textbf{Audio-visual results: }Performance results for visual-only, audio-only and audio-visual \methodName{} on the LRS2 test set with clean audio and in the presence of noise at 0~dB SNR. Standard deviations for this table are given in the appendix. Figure~4: (Right) Mean average precision for visual-only (red), audio-only (blue) and audio-visual (green) KWS-Net with a noisy audio signal, as the SNR is varied between -10 dB and 20 dB. } 
\vspace{-5mm}
\label{tab:audio_visual_main}

\end{table}

\vspace{-3mm}
 \subsection{Extension to other languages: French and German}

We now move on to assess the generalisation of our method to other languages.
For each of the experiments in Table~\ref{table:languages}, the model is first trained on LRW, then  fine-tuned on LRS2 and subsequently LRS3. For LRS3-Fr and LRS3-De, the
model is additionally fine-tuned on their corresponding training
set. Due to the lack of word timings for LRS3-Fr and LRS3-De, we train the models here without them (see loss description in Section~\ref{sec:model}). During evaluation, we do not consider the location of the maximum keyword detection probability. 

The more challenging setting of LRS3 compared to LRS2 (see Section~\ref{sec:exp_setup}) is reflected in the visual-only KWS; lip reading is also found to be harder on LRS3 compared to LRS2 \cite{Afouras19}. \lili{ In fact, we split the LRS3 test set into near-frontal and profile views: we find that the model is robust to side views (48.7\% mAP) but as expected, the performance is overall better on frontal clips (60.6\% mAP). }

The performance on LRS3-Fr is close to that on LRS3: the audio-only EER is slightly worse as a lot more English audio from LRW and LRS2 is used for training. The visual-only EER for LRS3-De is higher than LRS3-Fr (13.0\% vs.\ 8.4\%). However, LRS3-~Fr training set is five times bigger than that of LRS3-De (see Table~\ref{table:datasets}). In all cases, the audio-visual model performs better than audio-only and visual-only. The results in Table~\ref{table:languages} show that KWS-Net can be used for other languages, even if less language specific data is available.
 \begin{table}[!h]
 \vspace{-2mm}
\footnotesize
\centering

\setlength{\tabcolsep}{10pt}
\begin{tabularx}{\linewidth}{ c c c c c c c }
\toprule
 Dataset & Modality  &R@1* & R@5* & R@10* & mAP* & EER* \\
 \midrule
  \midrule
  LRS3& V & 25.5 $\pm$ 0.4 &50.0 $\pm$ 0.5 &62.1 $\pm$ 0.3 & 45.7 $\pm$ 0.3 & 8.3 $\pm$ 0.3 \\
 LRS3 & A & 52.0 $\pm$ 0.9 & 88.4 $\pm$ 0.5 & 94.0 $\pm$ 0.4& 85.2 $\pm$ 0.6 & 2.1 $\pm$ 0.1 \\
 LRS3& AV  & 55.4 $\pm$ 0.9 & 90.6 $\pm$ 0.2 & 95.9 $\pm$ 0.2& 88.3 $\pm$ 0.4& 1.6 $\pm$ 0.1  \\
\midrule
LRS3-Fr & V   & 28.8 $\pm$ 0.3  & 55.3 $\pm$ 0.9 &  65.8 $\pm$ 0.7& 43.9 $\pm$ 0.3& 8.4 $\pm$ 0.1\\
LRS3-Fr & A & 52.3 $\pm$ 0.6 & 86.9 $\pm$ 0.2& 92.7 $\pm$ 0.2& 72.6 $\pm$ 0.3 &3.4 $\pm$ 0.1 \\
LRS3-Fr & AV   & 53.3 $\pm$ 0.4  & 88.9 $\pm$ 0.2 & 93.9 $\pm$ 0.3 & 74.1 $\pm$ 0.3&3.2 $\pm$ 0.1\\
 \midrule
 LRS3-De& V&  13.3 $\pm$ 0.1  &33.7 $\pm$ 0.1 &43.5 $\pm$ 0.2& 24.9 $\pm$ 0.1& 13.0 $\pm$ 0.2 \\
LRS3-De & A  & 48.1 $\pm$ 0.4   &79.9 $\pm$ 0.5 &88.1 $\pm$ 0.2& 67.4 $\pm$ 0.3 & 3.7 $\pm$ 0.2  \\
LRS3-De & AV  & 50.5 $\pm$ 0.3  & 83.3 $\pm$ 0.1 &90.2 $\pm$ 0.1& 70.3 $\pm$ 0.2  & 3.4 $\pm$ 0.1\\
\bottomrule
\end{tabularx}
\vspace{-3mm}
\caption{\textbf{Language results}: Performance of visual-only, audio-only and audio-visual
\methodName{} on LRS3 (English), LRS3-Fr (French) and LRS3-De
(German). *The task here is classifying whether the keyword occurs in the clip, and 
keywords may be seen during training.
}
\vspace{-6mm}
\label{table:languages}
\end{table}

\section{Conclusion}

\noindent In this paper, we present a novel CNN-based KWS architecture, \methodName, inspired by object detection methods. Our best visual-only model exceeds the performance of the previous state of the art on the LRS2 dataset. We show that combining audio and visual modalities helps KWS for both clean and noisy audio. Finally, we demonstrate that KWS-Net generalises to languages other than English. 
\lili{In future work, we plan to improve KWS-Net by incorporating context of surrounding words.} \\
\\

\noindent{\bf Acknowledgements.} We thank G{\"u}l Varol and Olivia Wiles for their helpful comments. Funding for this research is provided by the UK EPSRC
CDT in Autonomous Intelligent Machines and Systems, 
the Oxford-Google DeepMind Graduate Scholarship, and the EPSRC 
Programme Grant Seebibyte EP/M013774/1.

\bibliography{shortstrings,mybib}

\clearpage
\title{APPENDIX}
\maketitle
\input{appendix.tex}
\end{document}

%% file: appendix.tex
\renewcommand{\thefigure}{A.\arabic{figure}}
\setcounter{figure}{0} 
\renewcommand{\thetable}{A.\arabic{table}}
\setcounter{table}{0} 

\appendix

\section{Appendix}
\subsection{KWS-Net Architecture} \label{sec:arc_details}
\begin{figure*}[h]
 \vspace{-5mm}
    \centering
    \includegraphics[width=\linewidth]{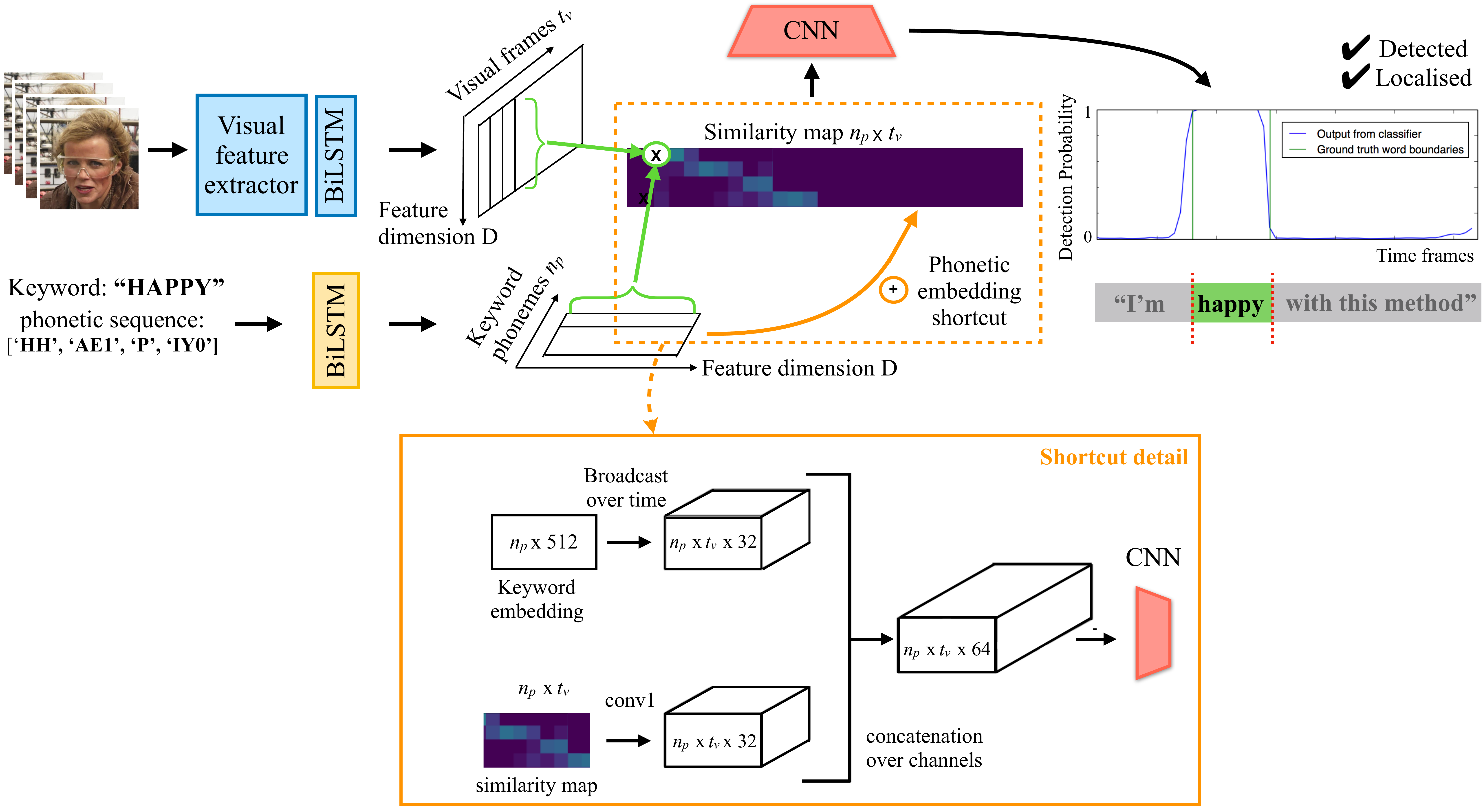}
    \vspace{-5mm}
    \caption{Architecture detail for Figure 2 of main paper.}
    \label{fig:novel_architecture}
\end{figure*}

\begin{table}[h!]

\centering
            \footnotesize
\begin{tabular}[t]{ l r r r r }
\multicolumn{5}{c}{\textbf{(a) Audio Feature Extractor}} \\
 \\
 \toprule
 Layer & \# filters & K & S   & Output  \\  
 \midrule
 input  & - &   - &  -                   &  $4t_v \times 80 \times 1$  \\  
 conv1    & 256 & 5 & 2   &  $2t_v \times 256 \times 1$\\  
 conv2 & 512 & 5   & 2     & $t_v \times 512 \times 1$ \\ 
 \bottomrule
\end{tabular}
\begin{tabular}[t]{  l@{} r r }
\multicolumn{3}{c}{\textbf{(b) Visual BiLSTM}} \\
 \\
  \toprule
 Layer & \# filters & Output  \\  
 \midrule
 input  & 512 &   $t_v \times 1 $  \\  
 BiLSTM    & 256 &  $t_v \times 1$ \\
 fc1   & 512 &  $t_v \times 1$ \\
 \bottomrule
 \vspace{3mm}
\end{tabular}
\begin{tabular}[t]{  l@{} r r }
\multicolumn{3}{c}{\textbf{(c) Phoneme BiLSTM}} \\
 \\
  \toprule
 Layer & \# filters & Output  \\  
 \midrule
 input  & 64 &   $n_p \times 1 $  \\  
 BiLSTM    & 500 &  $n_p \times 1$ \\
  fc1   & 128 &  $t_v \times 1$ \\
  fc2   & 512 &  $t_v \times 1$ \\
 \bottomrule
\end{tabular}

\begin{tabular}[t]{  l@{}  r r r  }

\multicolumn{4}{c}{\textbf{(d) Visual Feature Extractor \cite{themos2}}} \\
 \\
 \toprule
 Layer & \# filters & K & S   \\  
 \midrule
 conv1    & 64 & (5,7,7) & (1,2,2)   \\  
 mp & - &  (1,3,3)  & (1,2,2)     \\
 convblock1 & 64& $(3,3)\times 2$ & 1    \\ 
 convblock2 & 64 & $(3,3)\times 2$ & 1   \\
 convblock3 & 128 & $(3,3)\times 2$ & 2   \\ 
 convblock4 & 128 & $(3,3)\times 2$ & 1    \\
 convblock5 & 256 & $(3,3)\times 2$ & 2    \\ 
 convblock6 & 256 & $(3,3)\times 2$ & 1   \\
 convblock7 & 512 & $(3,3)\times 2$ & 2    \\ 
 convblock8 & 512 & $(3,3)\times 2$ & 1    \\
 
 \bottomrule
\end{tabular}
\hfill
\begin{tabular}[t]{  l r c c r }
\multicolumn{5}{c}{\textbf{(e) CNN}} \\
\\
 \toprule
 Layer &     \#filters & K & S   & Output  \\  
 \midrule
 input  & - &   - &  -             & $t_v \times  n_p \times 1$   \\
 
 conv1  & 32 & (5,5)   & (1,1)     & $t_v \times  n_p \times 32$ \\ 
 conv2   & 128   & (5,5)   & (2,2)     & $t_v/2 \times  n_p/2 \times 128$ \\ 
 mp1 & - & (2,2)   & (2,1)     &  $t_v/4 \times  n_p/2 \times 128$  \\ 
 conv3 & 256 & (5,5)   & (2,1)     & $t_v/8 \times  n_p/2 \times 256$ \\ 
 avgp1   & -   & (1,$n_p/2$)   & -     & $t_v/8 \times 1 \times 256 $\\ 
 fc1   & 512  & (1,1)   & (1,1)     & $t_v/8 \times 1 \times 512$ \\
 fc2   & 256 & (1,1)   & (1,1)     & $t_v/8 \times 1\times 256$  \\
 fc3   & 1 & (1,1)   & (1,1)     & $t_v/8 \times 1 \times 1$\\
 mp2   & -   & ($t_v/8$,1)   & -   & $1 \times 1 \times 1$ \\ 
 \bottomrule
\end{tabular}

\caption{\textbf{Architecture details for audio-only and visual-only KWS-Net}: $K$ denotes kernel width and $S$ the strides. \textit{mp} denotes a max-pooling layer. \textit{avgp} denotes an average-pooling layer. Batch Normalization and ReLU activation are added after every convolutional layer. \textit{$n_p$, $t_v$} denote the number of phonemes in the keyword and the number of frames in the video clip respectively. \textit{convblock} denotes a residual convolution block.}
\end{table}

\subsection{Baseline Architecture} \label{sec:baseline_arc}

\begin{figure*}[h]

    \centering
    \includegraphics[width=0.7\linewidth]{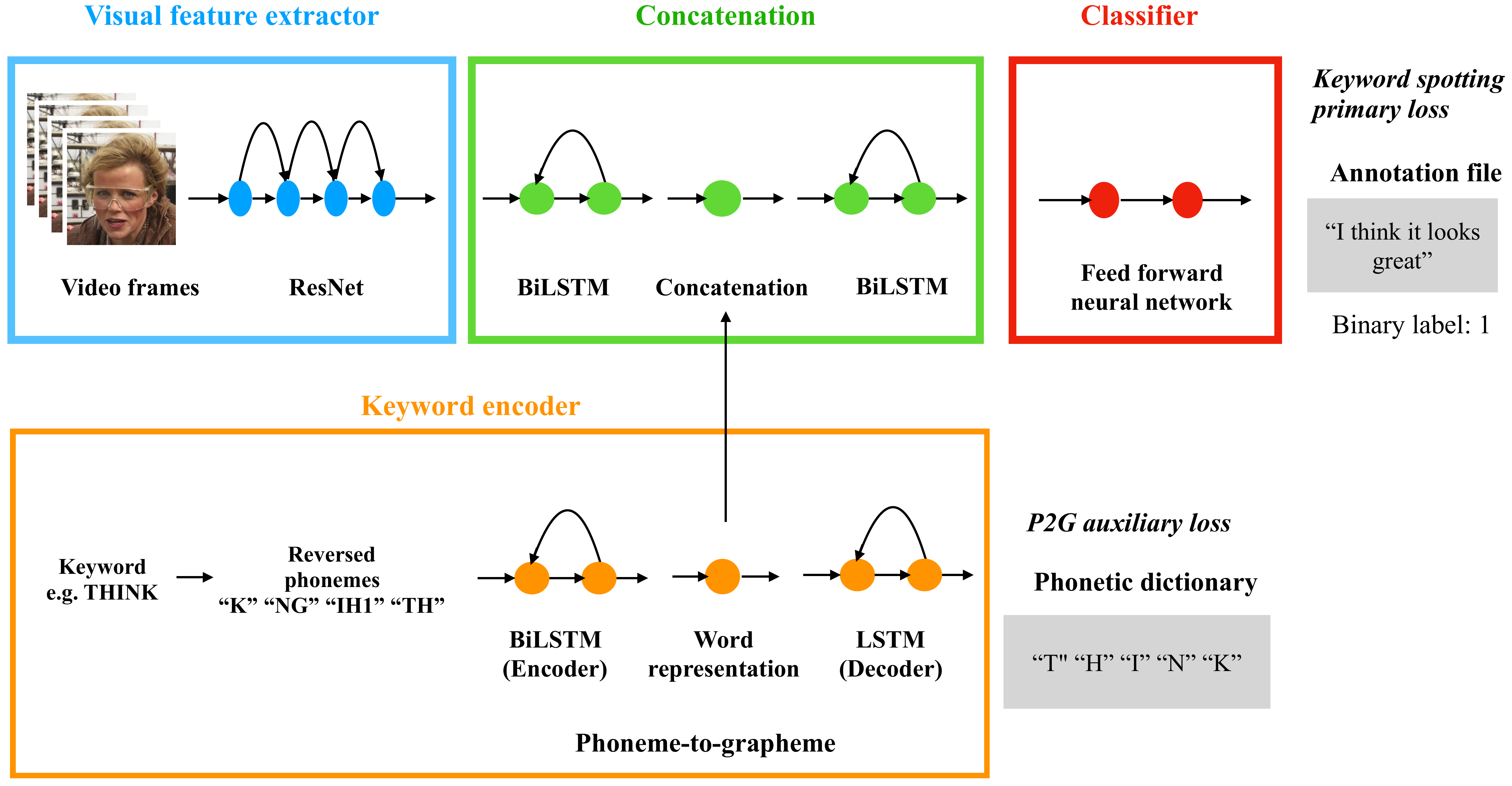}
    \caption{Architecture of baseline keyword spotting method.
    }

    \label{fig:baseline_arc}
\end{figure*}

The architecture of the baseline model of \cite{themoskws} that we use in the experiments is shown in Figure \ref{fig:baseline_arc}.
This method encodes the query keyword phoneme sequence into a compact embedding that is concatenated to the visual feature sequence. The concatenated feature is ingested by a BiLSTM followed by fully connected layers of a feed-forward network to output a binary prediction per frame. The model also contains a second output, a decoder that conditions on the phoneme embedding to predict the grapheme representation of the keyword and is trained with an auxiliary loss to provide regularisation. For more details please refer to \cite{themoskws}.
Note that the original paper inputs graphemes and outputs phonemes, whereas we have flipped this order, resulting in a P2G instead of a G2P model. We found this change to improve performance.

\vspace{-8mm}
\subsection{Datasets}
In Table~\ref{table:datasets_2} we give full details on the keyword vocabulary sizes used for training and testing for each dataset. For LRW and LRS2, the train-val and test set vocabularies are disjoint as we test in the zero-shot case with unseen keywords. For LRS3, LRS3-Fr and LRS3-De\footnote{\label{repeat}Both LRS3-Fr and LRS3-De are available from LRS3-Lang at \url{www.robots.ox.ac.uk/~vgg/data/lip_reading/lrs3-lang}}, this constraint is not imposed and we test with keywords seen and unseen during training.

\begin{table}
\footnotesize
 \centering
\begin{tabularx}{0.4\linewidth}{c c c }
\toprule
Dataset & Split & Vocabulary \\ 

\midrule
 \multirow{2}{*}{LRW~\cite{Chung16}}    & Train-val & 391  \\
                      & Test      & 109  \\
\midrule
 & Pre-train & 16k\\
LRS2~\cite{Chung16b, Afouras19} & Train-val & 8k \\
 & Test      & 644\\
 \midrule
 & Pre-train & 24k\\
LRS3~\cite{Afouras18d} & Train-val &  10k\\
 & Test      & 209\\
 \midrule
  \multirow{2}{*}{LRS3-Fr}    & Train-val & 17k \\
                      & Test      & 824  \\
\midrule
  \multirow{2}{*}{LRS3-De} & Train-val & 9k \\
                      & Test  & 1,135 \\
\bottomrule
\end{tabularx}

 \caption{Keyword vocabulary size used for each dataset split.}
 \label{table:datasets_2}

\end{table}

\subsection{Training curriculum}
\noindent
\textbf{Batch creation.} We follow the same procedure as \cite{themoskws} for batch creation. During an epoch, the clips are partitioned into mini-batches of size 40. For each batch, each clip is paired up with all its keywords to form positive examples and with an equal number of randomly chosen keywords, which are not uttered in the clip, to form negative examples. 

\noindent
\textbf{Optimisation.} The loss function is optimised with backpropagation using the Adam optimiser. The network is trained for 100 epochs (pre-training and fine-tuning) and the best model is chosen from the performance on the validation set. For pre-training on LRW, the initial learning rate is $10^{-3}$ and decreases by a half every 10 epochs. For fine-tuning on LRS2, the initial learning rate is $10^{-4}$ and decreases by a half every 20 epochs. The model is implemented in PyTorch.

\vspace{-3mm}
\subsection{Additional results} \label{sec:quant_results}
\noindent \textbf{Audio-visual KWS-Net.} We repeat Table 4 of the main paper with the error margins.

\begin{table}[h!]
\footnotesize
\centering
\begin{tabularx}{0.83\linewidth}{ c c c c c c c}
\toprule
Modality & Noise  & R@1 & R@5 & R@10& mAP & EER \\
 \midrule
  \midrule
V & \xmark  &37.9 $\pm$ 0.3 &66.8 $\pm$ 0.6&75.6 $\pm$ 0.5&53.9 $\pm$ 0.3 &5.7 $\pm$ 0.2\\
A   &\xmark    &67.7 $\pm$ 0.7 &91.1 $\pm$ 0.5& 94.6 $\pm$ 0.2&83.3 $\pm$ 0.5& 1.9 $\pm$ 0.2\\
 AV & \xmark &72.2 $\pm$ 0.4 &94.7 $\pm$ 0.4& 97.0 $\pm$ 0.4 &87.5 $\pm$ 0.4  & 1.7 $\pm$ 0.1\\
\midrule
A & \cmark  &27.6 $\pm$ 0.6 &49.8 $\pm$ 0.3& 59.4 $\pm$ 0.4&39.7 $\pm$ 0.5 & 12.8 $\pm$ 0.2 \\
 AV & \cmark  &52.7 $\pm$ 0.3 &81.9 $\pm$ 0.5& 87.0 $\pm$ 0.3 &69.6 $\pm$ 0.2& 4.3 $\pm$ 0.1 \\
\bottomrule
\end{tabularx}
\caption{Performance of visual-only, audio-only and audio-visual \methodName{} on the LRS2 test set for a clean audio signal and a noisy audio signal with SNR of 0 dB.} 
\label{tab:audio_visual}
\end{table}

\noindent\lili{\textbf{Late fusion modality weights.} We illustrate how varying the modality weights for late fusion impacts the model performance on the LRS3 test set in Table~\ref{tab:audio_visual_modality_weights}. As expected, the results improve when the audio modality is given more weighting for a clean audio signal. However, fusing with the video modality surpasses an audio-only model. The overall best performance is for audio weight  $w_{audio} = 0.7$ and video weight $w_{video} = 0.3$.}
\begin{table}[h!]
\footnotesize

\centering
\begin{tabularx}{0.6\linewidth}{ c c c c c c c}
\toprule
$w_{audio}$ & $w_{video}$  & R@1* & R@5* & R@10*& mAP* & EER* \\
 \midrule
1.0 & 0.0 & 51.6& 88.4& 93.9 &85.4& 2.1\\
0.9 & 0.1 & 52.6 & 89.5 &94.7&86.6&1.8\\
0.8 & 0.2 & 54.0 &89.8&95.5&87.8&1.8\\
\textbf{0.7} &\textbf{ 0.3 }&\textbf{ 57.1} & \textbf{90.7}&96.0&\textbf{89.0}& \textbf{1.4}\\
0.6 & 0.4 & 55.5 & 90.7&\textbf{96.5}&87.9&1.4 \\
0.5 & 0.5 &54.4 &89.3&96.3&86.9& 1.6\\
0.4 & 0.6 & 51.6 &89.6&95.2&84.5&1.8 \\
0.3 & 0.7 &48.4 &85.7&93.8&80.2& 2.5\\
0.2 & 0.8 & 44.9 &78.1&88.4&73.8&3.5 \\
0.1 & 0.9 & 35.8 &68.4&79.0&61.1&5.5 \\
0.0 & 1.0 &24.8 &49.5 &61.7&45.4&8.1 \\
\bottomrule
\end{tabularx}
\caption{Performance of audio-visual \methodName{} on the LRS3 test set with a clean audio signal for varying audio weighting $w_{audio}$ and video weighting $w_{video}$. *The task here is classifying whether the keyword occurs in the clip, and 
keywords may be seen during training.} 
\label{tab:audio_visual_modality_weights}
\end{table}

\clearpage
\subsection{Qualitative examples} \label{sec:qual_results}
Please visit our project webpage at \small{\url{www.robots.ox.ac.uk/~vgg/research/kws-net/}} \normalsize to see a video of qualitative results of our model in action. Examples are also shown below. 
\begin{figure*}[h!]
    \centering
       
    \includegraphics[width=0.9\linewidth]{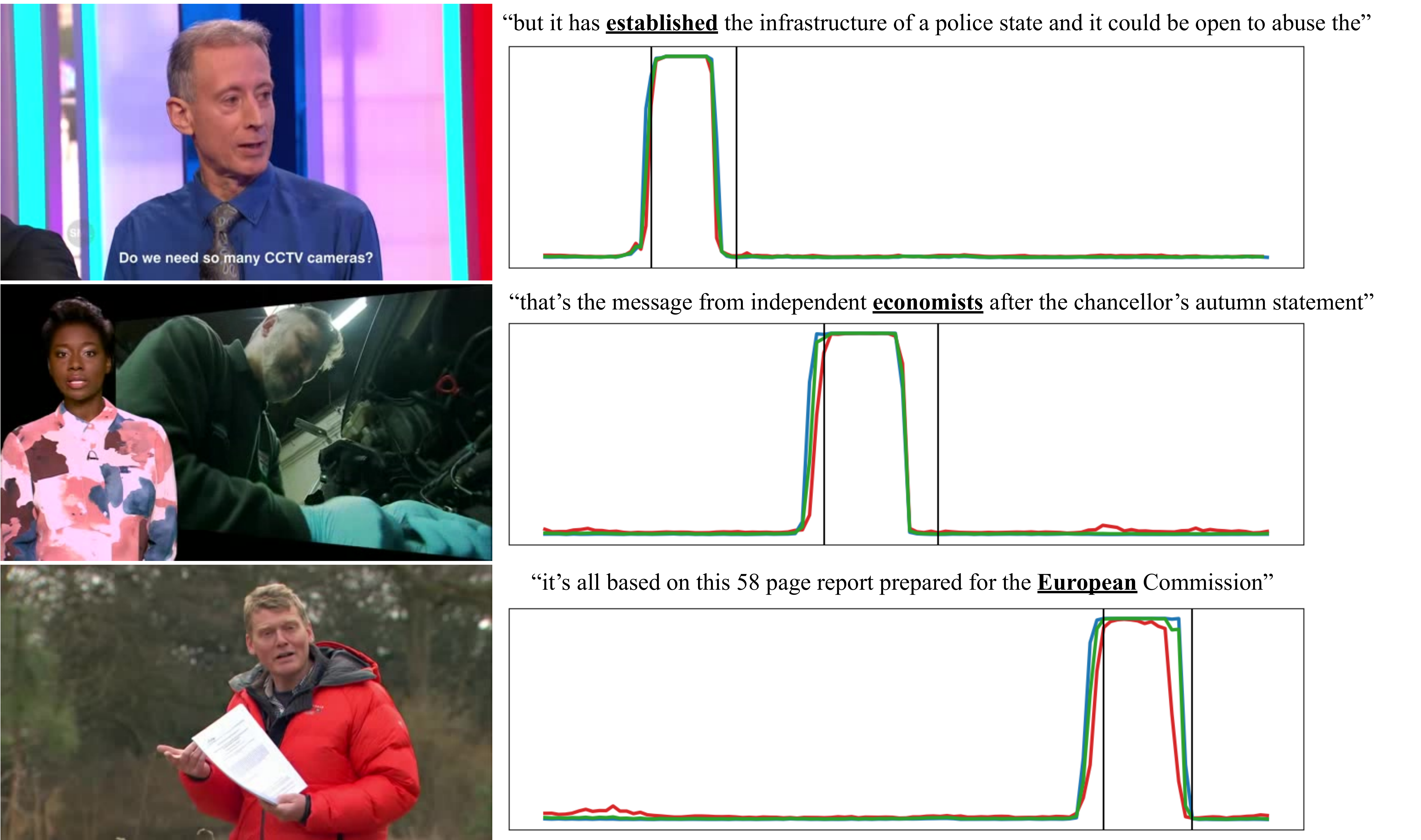}
    \includegraphics[width=0.9\linewidth]{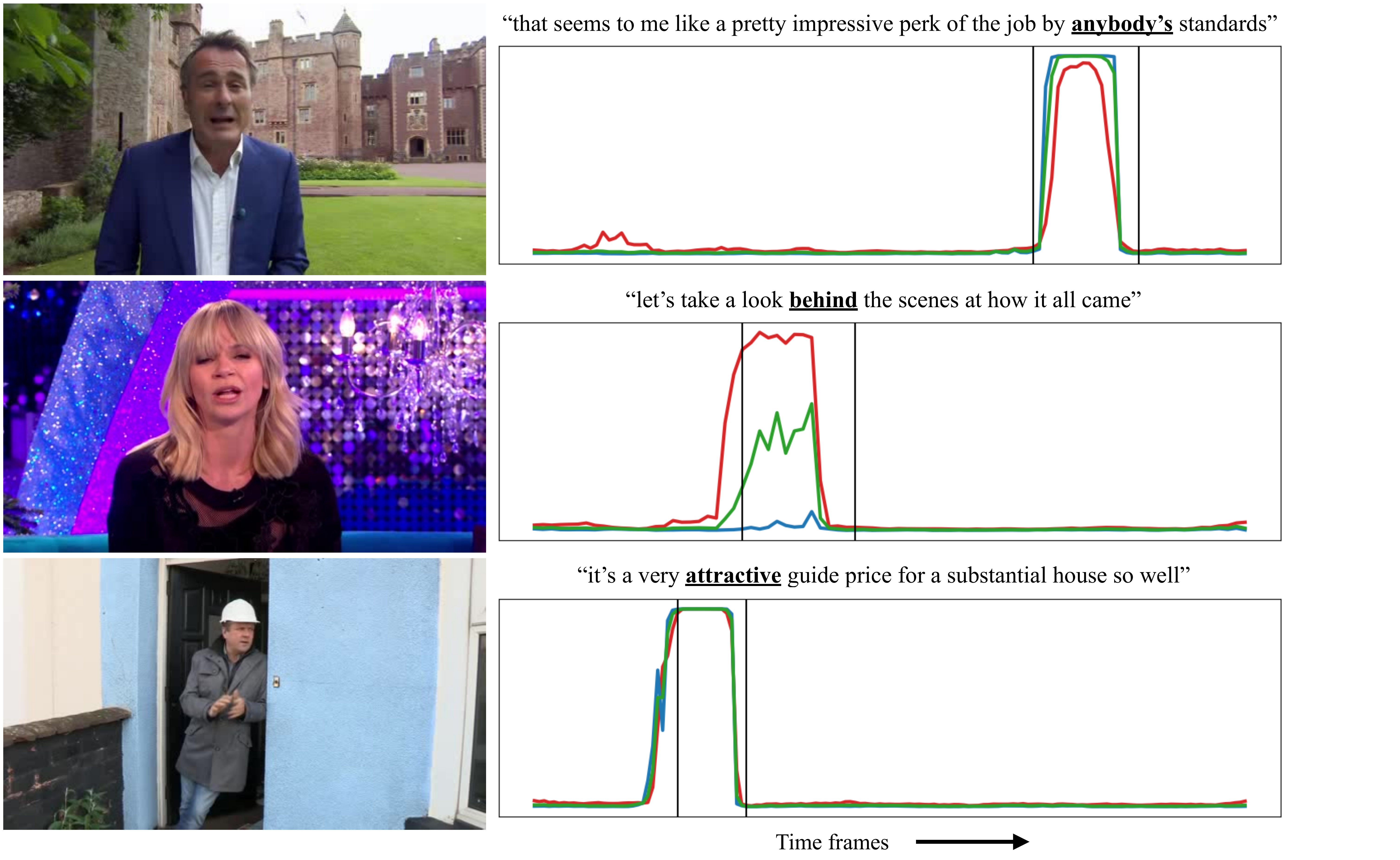}
 
 \caption{(Right column) Plots of keyword detection probability over time frames for visual-only (red), audio-only (blue) and audio-visual (green) KWS-Net for positive keyword-clip pairs in the LRS2 test set. The corresponding clip utterance is shown above each plot, with the user-defined keyword in bold and underlined. The vertical lines correspond to the ground truth start and end times of the user-defined keyword. (Left column) Frame extracted from corresponding LRS2 test set clip.}
\end{figure*}